\documentclass[journal,letterpaper]{IEEEtran}
\usepackage[pdftex]{graphicx}
\usepackage[numbers]{natbib}
\usepackage{booktabs}
\usepackage{moreverb}
\usepackage{url}
\usepackage{amsmath}
\usepackage{amsfonts}
\usepackage{bm}
\usepackage{subfig}
\usepackage[colorlinks,bookmarksopen,bookmarksnumbered,citecolor=red,urlcolor=red]{hyperref}
\hypersetup
{
	pdftitle = {Whole-body control for quadrupedal locomotion on challenging terrain},
	pdfauthor = {Shamel Fahmi and Carlos Mastalli},
	pdfsubject = {ArXiV manuscript},
	pdfkeywords = {whole-body control, legged robots, challenging locomotion, and
	terrain mapping},
	pdftoolbar = true,
	colorlinks = true,
	linkcolor = black,
	citecolor = black,
	urlcolor = black,
}
\usepackage{color}
\usepackage[usenames,dvipsnames]{xcolor}
\definecolor{blue_iit}{RGB}{51,51,255}
\usepackage{algpseudocode}
\usepackage{algorithm}
\usepackage{glossaries}
	\newacronym{hyq}{HyQ}{Hydraulically actuated Quadruped}
	\newacronym{lf}{LF}{Left-Front}
	\newacronym{rf}{RF}{Right-Front}
	\newacronym{lh}{LH}{Left-Hind}
	\newacronym{rh}{RH}{Right-Hind}
	\newacronym{ptu}{PTU}{Pan and Tilt Unit}
	\newacronym{imu}{IMU}{Inertial Measurement Unit}
	\newacronym{dofs}{DoFs}{Degrees of Freedom}
	\newacronym{rt}{RT}{Real Time}
	\newacronym{com}{CoM}{Center of Mass}
	\newacronym{haa}{HAA}{Hip Adduction-Abduction}
	\newacronym{hfe}{HFE}{Hip Flexion-Extension}
	\newacronym{kfe}{KFE}{Knee Flexion-Extension}
	\newacronym{cop}{CoP}{Center of Pressure}
	\newacronym{zmp}{ZMP}{Zero Moment Point}
	\newacronym{icp}{ICP}{Instantaneous Capture Point}
	\newacronym{cmp}{CMP}{Centroidal Moment Pivot}
	\newacronym{grfs}{GRFs}{Ground Reaction Forces}
	\newacronym{mcot}{MCoT}{Mechanical Cost of Transport}
	\newacronym{cmm}{CMM}{Centroidal Momentum Matrix}
	\newacronym{rnea}{RNEA}{Recursive Newton-Euler Algorithm}
	\newacronym{slip}{SLIP}{Spring Loaded Inverted Pendulum}
	\newacronym{eom}{EoM}{Equation of Motions}
	\newacronym{aras}{ARA$^*$}{Anytime Repairing A$^*$}
	\newacronym{mpc}{MPC}{Model Predictive Control}
	\newacronym{qp}{QP}{Quadratic Program}
	\newacronym{sqp}{SQP}{Sequential Quadratic Programming}
	\newacronym{mic}{MIC}{Mixed-Integer Convex}
	\newacronym{cmaes}{CMA-ES}{Covariance Matrix Adaptation Evolution Strategy}
	\newacronym{ara}{ARA*}{Anytime Repairing A*}
	\newacronym{pca}{PCA}{Principal Component Analysis}
	\newacronym{cpg}{CPG}{Central Pattern Generator}
	\newacronym{wbc}{WBC}{Whole-Body Control}
\usepackage[tight]{units}
\usepackage[normalem]{ulem} 
\newcommand{\sref}[1]{Section~\ref{#1}}

\newcommand{\eref}[1]{(\ref{#1})}

\newcommand{\fref}[1]{Fig.~\ref{#1}}

\usepackage{cleveref}
\crefname{figure}{Fig.}{Fig.}
\crefname{equation}{Eq.}{Eq.}
\AtBeginDocument{%
}
\newtheorem{assump}{Assumption}
\newtheorem{assumpB}{Assumption}

\newcommand{\assref}[1]{Assumption~\ref{#1}}
\newcommand{\reducespace}{\vspace{-1.5em}}
\newcommand{\Rnum}{\mathbb{R}} 

\newcommand{\grf}{F_{\mathrm{grf}}} 
\newcommand{\mrm}[1]{\mathrm{#1}}
\newcommand{\nmrm}[1]{{#1}}

\newcommand{\vc}[1]{#1}
\DeclareMathOperator*{\st}{s.t.}						
\newcommand{\mat}[1]{\ensuremath{\begin{bmatrix}#1\end{bmatrix}}}	

\newcommand\BibTeX{{\rmfamily B\kern-.05em \textsc{i\kern-.025em b}\kern-.08em
T\kern-.1667em\lower.7ex\hbox{E}\kern-.125emX}}
\makeglossaries

\title{\textcolor{black}{Passive Whole-body Control for Quadruped Robots: \\ Experimental Validation over 
Challenging 
Terrain}}
\author{Shamel Fahmi$^{1\dagger}$, Carlos Mastalli$^{1,2\dagger}$, Michele
Focchi$^1$ and  Claudio Semini$^1$
\thanks{$^\dagger$These authors contributed equally to this work.}
\thanks{$^1$Dynamic Legged Systems Lab, Istituto Italiano di Tecnologia (IIT),
Genova, Italy.
{\tt\small \href{mailto:shamel.fahmi@iit.it}{firstname.lastname@iit.it}}}
\thanks{$^2$Gepetto Team, LAAS-CNRS, Toulouse, France.
{\tt\small \href{mailto:carlos.mastalli@laas.fr}{carlos.mastalli@laas.fr}}.}}

\usepackage{pdfpages}

\begin{document}
\null
\includepdf[pages=-]{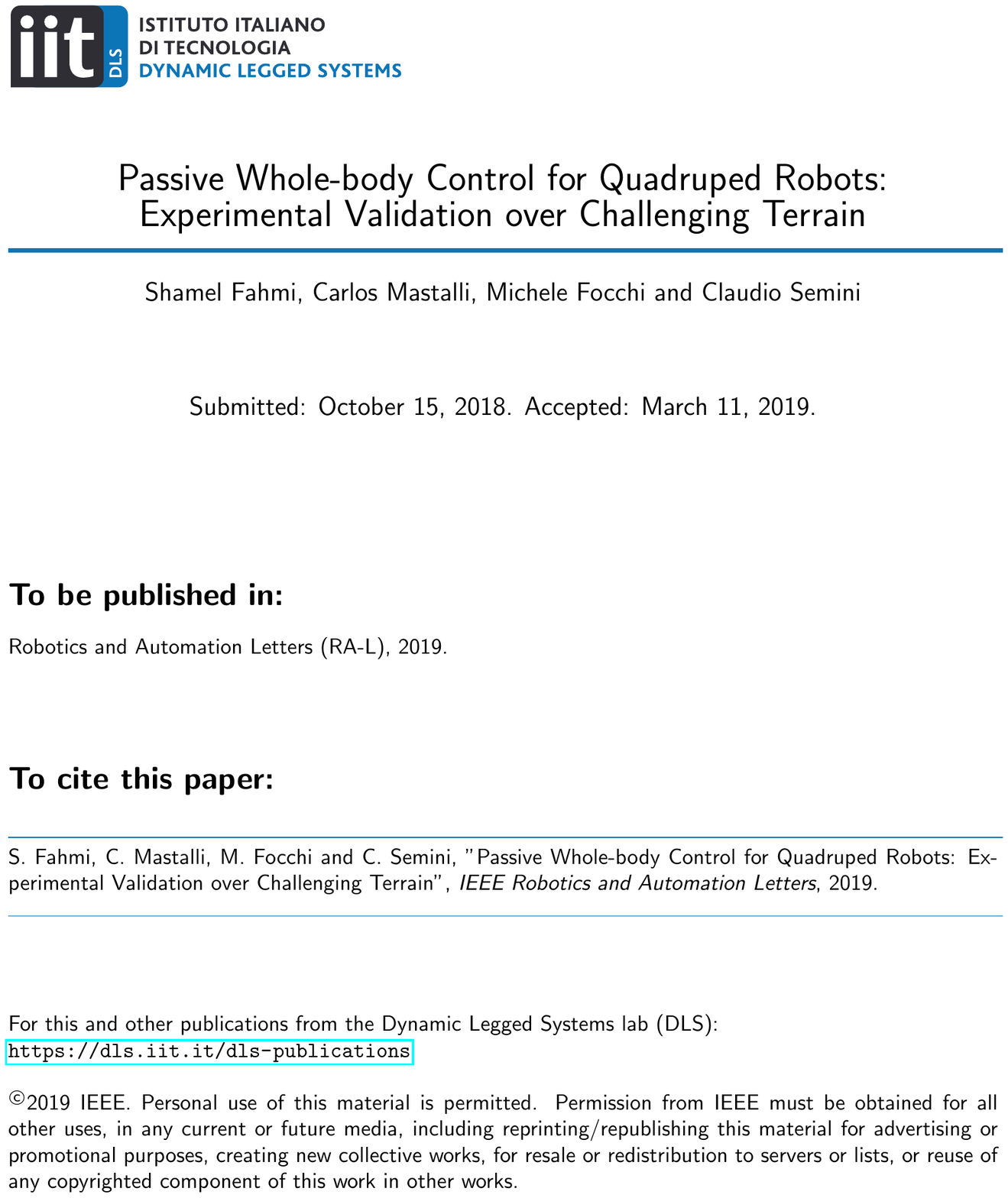}

\maketitle
\thispagestyle{empty}
\pagestyle{empty}

\begin{abstract}
We present \textcolor{black}{experimental results using} a passive \textcolor{black}{whole-body control} approach for 
quadruped robots
that achieves \textit{dynamic} locomotion while compliantly balancing the robot's trunk.
We formulate the motion tracking as a \gls{qp}  that takes into account
the full robot rigid body dynamics, the actuation limits, the joint limits 
and the contact interaction. 
We analyze the controller\textcolor{black}{'s} robustness against
inaccurate friction coefficient estimates and unstable footholds, 
as well as its capability to redistribute the load  
as a consequence of enforcing actuation limits.
Additionally, we \textcolor{black}{present practical} implementation details gained
from the experience with the real platform.
Extensive experimental trials on the \unit[90]{\textcolor{black}{kg}} \gls{hyq} robot validate the
capabilities of this controller under various terrain conditions and gaits.
The proposed approach is \textcolor{black}{superior} for accurate 
execution of high\textcolor{black}{ly} dynamic motions with respect 
to the current state of the art. 

\end{abstract}
\section{Introduction}\label{sec:introduction}
Achieving dynamic locomotion 
requires reasoning about the robot's dynamics, actuation limits and interaction with the environment 
while traversing challenging terrain (such as rough
or sloped \textcolor{black}{terrain).}
Optimization-based techniques can be exploited to  attain these objectives
in locomotion planning and control of legged robots.
For instance, one approach is to \textcolor{black}{use non-linear} \gls{mpc} 
while taking into consideration the full \textcolor{black}{dynamics of the} robot. 
Yet, 
it is often challenging to meet real-time
requirements because the solver can  get stuck in local minima, 
unless  proper warm-starting  is used \cite{Erez2013}.
Thus, 
current research often \textcolor{black}{relies} on low dimensional models or constraint relaxation
approaches to meet such requirements \textcolor{black}{(e.g. \cite{Kuindersma2014})}. 
Other approaches rely on decoupling the motion planning  from the  motion control
\cite{Farshidian2017b,Cabezas2018,Bellicoso2018}. 
Along this line, an optimization-based  motion planner 
could rely on low dimensional models to compute 
\gls{com} trajectories and footholds while a locomotion controller  \textcolor{black}{tracks} these
trajectories. 

Many recent contributions in locomotion control have been proposed in the
literature that \textcolor{black}{were successfully tested on} bipeds and quadrupeds
(e.g. \cite{Herzog2016,Koolen2016,Henze2016,Farshidian2017a,Bellicoso2018,Kim2018}). Some of them
are based on  quasi-static
assumptions or lower dimensional models  \cite{Stephens2010,Ott2011,focchi2017auro}.
This often limits the dynamic locomotion capabilities of the robot \cite{Herzog2016}. 
\textcolor{black}{Consequently, another approach, that is preferable for dynamic motion, is based on
\gls{wbc}. \gls{wbc} facilitates such decoupling between the motion planning and control in such a way that it is easy 
to accomplish multiple tasks while respecting the robot's behavior \cite{Farshidian2017a}.
These tasks might include motion tasks for the robot's end effectors 
(legs and feet) \cite{Henze2016,Farshidian2017a}, but also could be 
utilized for contacts anywhere on the robot's body \cite{Henze2017} or for a 
cooperative manipulation task between robots \cite{Bouyarmane2018}. }
%
%
%
%
\gls{wbc} casts the locomotion controller 
as an optimization problem,  in which, by incorporating the full dynamics of the legged robot, 
all of its \gls{dofs} are exploited in order to spread the desired motion tasks globally to all the joints.
This allows us to reason about multiple tasks  and solve them in an optimization fashion while respecting the 
full system dynamics and the actuation and interaction constraints. 
%
%
%
\gls{wbc} relies on the fact that robot dynamics 
and constraints could be formulated, at each loop, as linear constraints via \gls{qp}  \cite{Kuindersma2014}.
This allows us to solve the optimization problem in real-time.

%
Passivity theory is proven to guarantee a certain degree of robustness during interaction with the environment 
\cite{Stramigioli2015}. For that reason, such tool is commonly exploited in the design of locomotion controllers to 
ensure a 
passive contact interaction.  
Passivity based \gls{wbc} in humanoids was introduced first by \cite{Hyon2007} to effectively balance 
the robot when experiencing contacts. By providing compliant tracking and gravity compensation, the humanoid was 
able 
to adapt to unknown disturbances. The same approach was further extended first  by \cite{Ott2011} and later by 
\cite{Henze2016}. 
The former extended \cite{Hyon2007}  to posture control, 
while the  latter analyzed the passivity of a humanoid 
robot in multi-contact scenarios (by exploiting  the 
similarity with PD+ control \cite{Ortega2013}).
\begin{figure*}[tb]
	\centering
	\includegraphics[width=1.0\textwidth]{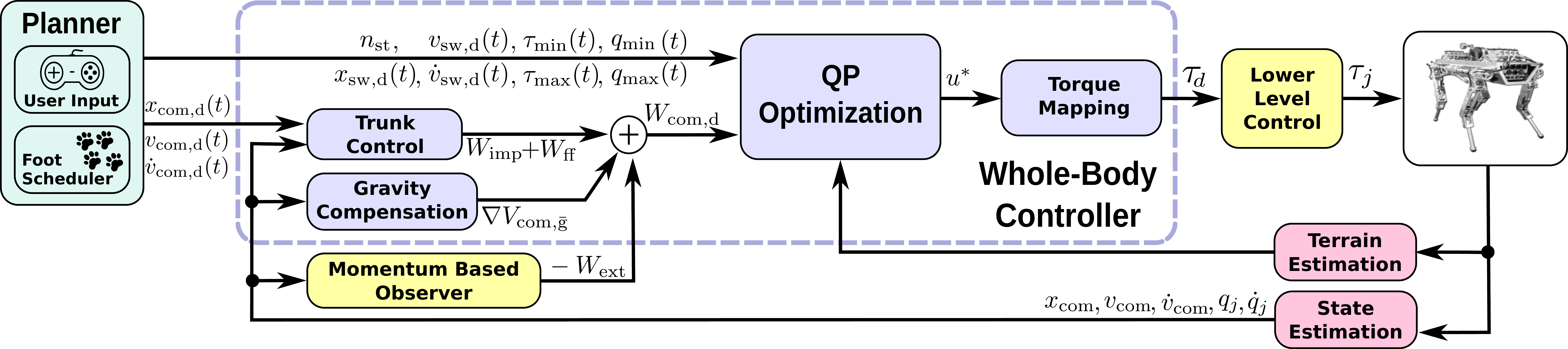}
	\reducespace
	\caption{\small \textcolor{black}{Overview of the whole-body controller as part of our
		locomotion framework}. 
	}
	\label{fig:blockDiagram}
\end{figure*}

In our previous work \citep{focchi2017auro}, 
the locomotion controller was
designed for quasi-static motions using only the robot's centroidal dynamics. 
Under that assumption, we noticed that 
during dynamic motions, the effect of the leg dynamics is no longer negligible; and \textcolor{black}{thus}, it
becomes necessary to abandon the quasi-static assumption to achieve  good tracking.
Second, since the robot is constantly interacting  with the
environment (especially during walking and running), it is crucial to ensure a
compliant and passive interaction.
For these reasons, in this paper, 
we improve our \textcolor{black}{previous work} \cite{focchi2017auro} by implementing a passivity based \gls{wbc} 
that incorporates the full robot dynamics  and interacts  compliantly with the 
environment, while satisfying the kinematic and torque limits.
Our \gls{wbc} implementation is capable of achieving faster dynamic motions 
than our \textcolor{black}{ previous work}.
%
%
We also integrate terrain mapping and state estimation on-board 
and present some practical implementation details  gained from the experience with the real platform. 

\textit{Contributions:} 
In this paper, we mainly present \textit{experimental} contributions in which we 
demonstrate the effectiveness of the controller both in simulation and experiments on \gls{hyq}. 
%
%
%
%
Compared to previous work on passivity-based \gls{wbc} \cite{Henze2016,Ott2011},  
in which experiments were conducted on the robot \textcolor{black}{while} standing (not walking or running),
we tested our controller on \gls{hyq} during crawling and trotting. 
%
%
\textcolor{black}{Similar to the recent successful work of  \cite{Bellicoso2018} and \cite{DiCarlo2018}
 in  quadrupedal 
locomotion over rough terrain, we used similar terrain templates to present experiments of our passive \gls{wbc} on 
\gls{hyq} using multiple gaits over slopes and rough terrain of different heights.}

The rest of this paper is structured as follows:
%
In \sref{sec:wholeBodyController} we present the detailed formulation and design of our \gls{wbc}
followed by its passivity analysis in \sref{sec:Passivity}. 
\sref{sec:Implementations} presents further crucial implementation details. 
Finally we present our simulation and experimental results in \sref{sec:expResults} followed by our conclusions in 
\sref{sec:conclusion}. 
\section{Whole-body controller (WBC)}
\label{sec:wholeBodyController}
In this section we present and formulate our \gls{wbc}. 
%
%
Figure \ref{fig:blockDiagram} depicts the main components of our locomotion framework. 
Given high-level commands, the planner generates a reference motion
online \cite{Focchi2018star} or offline \cite{Cabezas2018}, and provides it to the 
\gls{wbc}.
Such references include the desired trajectories for \gls{com}, trunk orientation
and  swing legs. 
The state estimation supplies the controller with an estimate of the actual state of the robot,
by fusing  leg odometry, inertial sensing, visual odometry and LIDAR
\cite{nobili2017a} 
while, the 
terrain estimator, provides an estimate of the  terrain inclination (i.e. surface normal). 
Finally, there is a momentum-based observer that estimates external 
disturbances \cite{Focchi2018star} and a torque controller.

The goal of the designed \gls{wbc} is to keep the quadruped robot balanced (during running, walking or 
standing) while interacting passively with the environment.
The motion tasks of a quadruped robot can be categorized into 
a \textit{trunk task} and a \textit{swing task}. 
The trunk task regulates the position  of the \gls{com} and the  orientation of the trunk\footnote{Since \gls{hyq} is 
not equipped with arms, it suffices for us to control the trunk orientation instead of 
the whole robot angular momentum.}  and is achieved by 
implementing a Cartesian-based impedance controller with a feed-forward term\footnote{This is similar to a PD+ 
controller \cite{Ortega2013}.}.
The swing task regulates the swing foot trajectory in order to place 
it in the desired location while achieving enough clearance from the terrain.
Similar to the trunk task, the swing task is achieved by implementing a Cartesian-based impedance controller with a 
feed-forward term. 
The \gls{wbc} realizes these tasks by computing the optimal generalized accelerations and contact forces 
\cite{Herzog2016} via \gls{qp} and mapping them to the desired 
joint torques while taking into 
account the full dynamics of the robot, the properties of the terrain (\textit{friction constraints}), 
the unilaterality of the contacts (e.g. the leg can only push and not pull) (\textit{unilateral constraints}), 
and the actuator's \textit{torque/kinematic limits}. 
The desired torques,  will be sent to a 
lower-level (torque) controller.
%
%
%
%
%
%
%
%
\vspace{-0.5cm}
\subsection{Robot Model} 
\label{secRobotModel}

\begin{figure}[tb]
	\centering
	\includegraphics[width=0.3\textwidth]{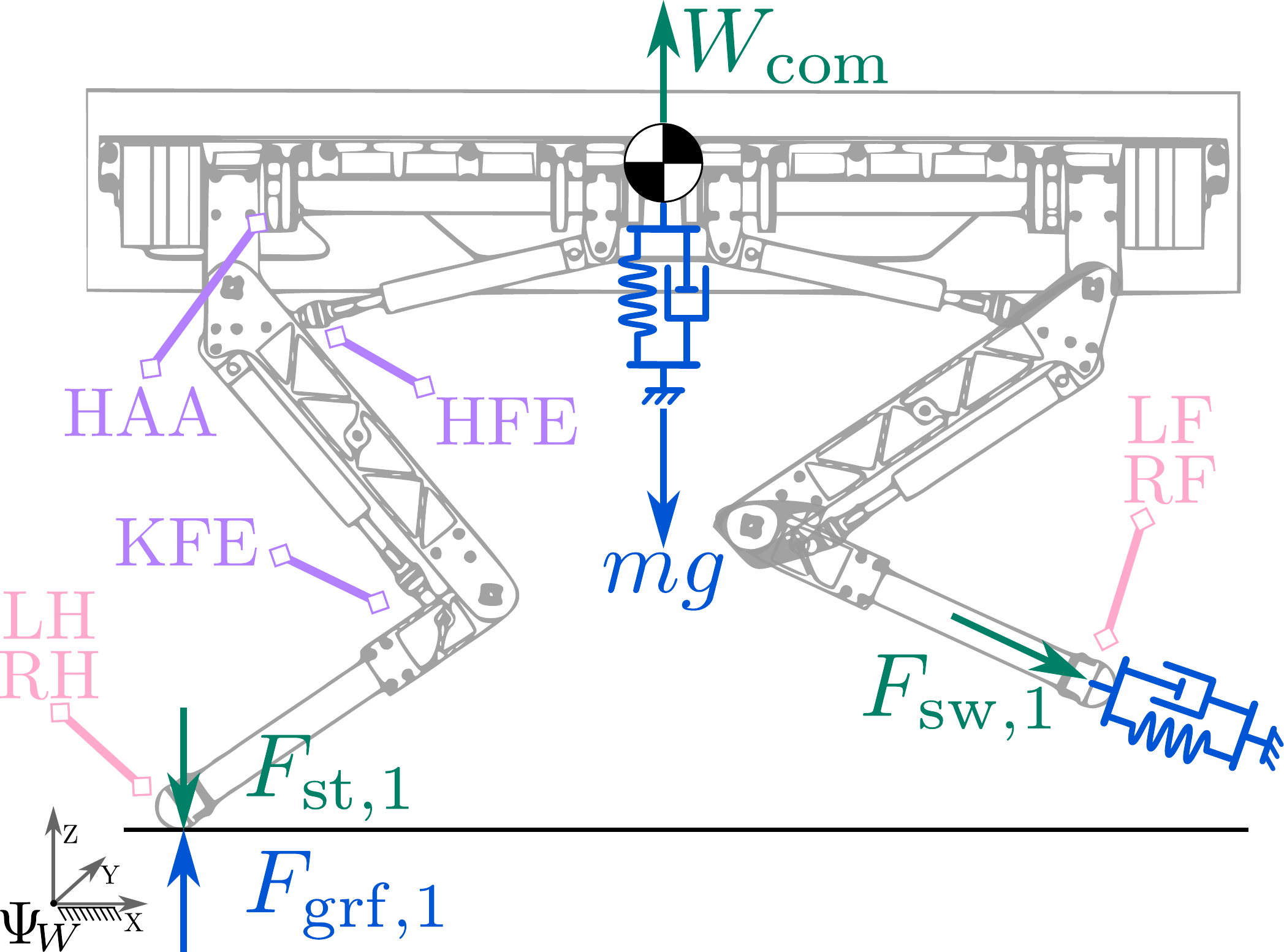}
	\caption{\small \textcolor{black}{Overview of the robot's joints, feet and generated wrenches and forces.
		LH, LF, RH and RF are Left-Hind, Left-Front, Right-Hind and Righ-Front legs, respectively. HAA, HFE and KFE are 
		Hip Abduction/Adduction, Hip Flexion/Extension and Knee Flexion/Extension, respectively.}}
	\label{fig:conventions}
	\vspace{-0.3cm}
\end{figure}

For a legged robot with $n$ \gls{dofs} and $c$  feet,
the forward kinematics of each foot is defined by $n_{\nmrm{a}}$  coordinates\footnote{Without the loss of 
generality, we consider a quadruped robot with $n=12$ \gls{dofs} with point feet, where $c = 4$ and 
$n_{\nmrm{a}} = 3$.}.
The total dimension of the feet operational space is $n_{\nmrm{f}} = n_{\nmrm{a}} 
c$. This can be separated into stance  ($n_{\mrm{st}} = n_{\nmrm{a}} c_{\mrm{st}}$) and swing feet 
($n_{\mrm{sw}} = n_{\nmrm{a}} c_{\mrm{sw}}$).
Since we are interested in regulating the position of 
the \gls{com},  we formulate the dynamic 
model in terms of the \gls{com}, using its
velocity rather than the base velocity\footnote{In this coordinate 
system, the inertia matrix is block diagonal \citep{Ott2011}. For the detailed 
implementation of the dynamics using the base velocity, see \cite{Hyon2007}.}
\citep{Ott2011}. Assuming that all the external forces are exerted on the \textit{stance feet}, we write the equation of motion that 
describes the full dynamics of the robot as:
\begin{equation}
\hspace{-0.5em}
\underbrace{\mat{M_{\mrm{com}} &	\vc{0}_{6\times n} \\
		\vc{0}_{n\times 6} & \bar{\vc{M}}_j}}_{\vc{M}(\vc{q})} \!\!
\underbrace{\mat{\dot{\vc{v}}_\mrm{com} \\ \ddot{\vc{q}}_j}}_{\ddot{\vc{q}}}\! + \! 
\underbrace{\mat{h_{com} \\ \bar{\vc{h}}_j}}_{h} \! = \!
\mat{ \vc{0}_{6\times n} \\ \vc{\tau_j}} +
\underbrace{\mat{\vc{J}_\mrm{st,com}^T \\ \vc{J}_\mrm{st,j}^T}}_{\vc{J}_\mrm{st}(\vc{q})^T}\vc{\grf},
\hspace{-0.5em}
\label{eq:full_dynamicsCOM}
\end{equation}
where the first 6 rows  represent the (un-actuated) floating 
base part and the remaining $n$ rows represent the actuated part.
$\vc{q}  \in SE(3) \times \Rnum^n$ represents the pose of the whole floating-base system while
$\vc{\dot{q}}=\mat{v_\mrm{com}^T & \vc{\dot{q}}_j^T}^T \in \Rnum^{6+n}$  and  $\ddot{\vc{q}} = [\dot{v}_{\mrm{com}}^T \ 
\ddot{q}_j^T]^T\in\Rnum^{6+n}$  are the vectors 
of generalized velocities and accelerations, respectively.
$v_\mrm{com} = [\dot{x}_\mrm{com}^T \  \vc{\omega}_b^T]^T$ $\in \Rnum^{6}$ 
and $\dot{v}_\mrm{com} = [\ddot{x}_\mrm{com}^T \  \dot{\vc{\omega}}_b^T]^T$ $\in \Rnum^{6}$
are the spatial velocity and acceleration of the floating-base expressed at the \gls{com}.
$\vc{M}(q) \in \Rnum^{(6+n) \times (6+n)}$ is the  inertia matrix,
where $M_\mrm{com}(q)$ $\in \Rnum^{6\times6}$ is the 
composite rigid body inertia matrix of the robot expressed at the \gls{com}.
$\vc{h}\in \Rnum^{6+n}$ is the force vector that accounts for Coriolis,
centrifugal, and gravitational forces
\footnote{Note that $h_{com} = -mg + 
v_{com} \times^*M_{com} v_{com}$ 
according to the spatial algebra notation, where $m$ is the total robot mass.}.
%
%
%
%
%
%
$\vc{\tau}\in\Rnum^n$ are the actuated joint torques while
$\vc{\grf} \in\Rnum^{n_\mrm{st}}$ is the vector of \gls{grfs} (contact forces).
%
%
In this context, the floating base Jacobian $J$ $\in\Rnum^{n_\mrm{f} \times(6+n)}$
is separated into swing Jacobian $J_{\mrm{sw}}$  $\in\Rnum^{n_{\mathrm{sw}} \times (6+n)}$ and stance 
Jacobian  $J_{\mrm{st}}$   $\in\Rnum^{n_{\mathrm{st}} \times (6+n)}$ which could be further expanded into 
$J_{\mathrm{st,com}}$  $\in\Rnum^{n_{\mathrm{st}} \times 6}$, $J_{\mathrm{st,j}}$  $\in\Rnum^{n_\mathrm{st} \times n}$, 
$J_{\mathrm{sw,com}}$ $\in\Rnum^{n_\mathrm{sw} \times 6}$ and  $J_{\mathrm{sw,j}}$ $\in\Rnum^{n_\mathrm{sw} \times 
n}$.  
%
%
%
%
The  operator $\bar{[\cdot]}$ denotes  the
matrices/vectors recomputed after the coordinate transform to the \gls{com} \cite{Hyon2007}. 
Following the sign convention in \fref{fig:conventions}, recalling the first 6 rows in 
\eref{eq:full_dynamicsCOM}, and  by defining the gravito-inertial  \textit{\gls{com} wrench} as $W_\mrm{com} = M_\mrm{com}
\dot{v}_\mrm{com} +  h_{com} $ $\in \Rnum^6$, we can write the \textit{floating-base dynamics} as:
\begin{eqnarray}
	W_\mrm{com}  =  J_\mrm{st,com}^T \grf ,
\label{eqn4}
\end{eqnarray}
such that  $J_{\mrm{st,com}}$ maps $\grf$ to the \gls{com} wrench space.

	The feet velocities $v$ $\in\Rnum^{n_\mrm{f}}$ could be separated into stance $v_{\mrm{st}}$ 
	$\in\Rnum^{n_{\mrm{st}}}$ and swing $v_{\mrm{sw}}$ $\in\Rnum^{n_{\mrm{sw}}}$ feet velocities.
	The mapping between $v= [v_{\mrm{st}}^T \ v_{\mrm{sw}}^T]^T$ and the generalized velocities $\dot{q}$ is: 
	\begin{subequations}
		\begin{eqnarray}
		v &=& J \dot{q} ,
		\label{pass1} \\
		v  =\begin{bmatrix} J_{\mrm{com}} & J_{\nmrm{j}} \end{bmatrix}\begin{bmatrix} v_\mrm{com}  \\ 
		\dot{q}_\nmrm{j} \end{bmatrix}
		&=&  J_{\mrm{com}}v_\mrm{com} +  J_{\nmrm{j}} \dot{q}_\nmrm{j},
		\label{eq_pass4}
		\end{eqnarray} 
	\end{subequations}
such that  $J_{\mathrm{com}}$  $\in\Rnum^{n_{\mathrm{f}} \times 6}$  and $J_{\nmrm{j}}$ $\in\Rnum^{n_{\mathrm{f}} 
\times n}$. Similar to the feet velocities, we split the feet force vector $F = [F_{\mrm{st}}^T \ 
F_{\mrm{sw}}^T]^T$ 
$\in\Rnum^{n_\mrm{f}}$, into  $F_{\mrm{st}}$ $\in\Rnum^{n_\mrm{st}}$ and $F_{\mrm{sw}}$  $\in\Rnum^{n_\mrm{sw}}$.
\begin{assump}
	The robot is walking over rigid terrain in which the 
	stance feet do 
	not move (i.e., $v_{\mrm{st}}= 
	\dot{v}_{\mrm{st}} = 0$).
	\label{ass1}
\end{assump}
\vspace{-0.25cm}
\subsection{Trunk and Swing Leg Control Tasks}
\label{secTrunkTask}
To compliantly achieve a desired motion of the \textit{trunk}, 
we define the desired wrench at the CoM  $W_{\mrm{com,d}}$ 
using the following: 
1) a Cartesian impedance at the \gls{com} $W_\mrm{imp}$ that is represented by
a stiffness term  ($\nabla V_{\mrm{com,K}} = K_{\mrm{com}}\Delta x_{\mrm{com}}$)  with positive definite stiffness 
matrix $K_{\mrm{com}}\in R^{6\times6}$
and a damping term ($D_{\mrm{com}} \Delta v_{\mrm{com}}$) with positive definite damping matrix $D_{\mrm{com}} \in R^{6\times6}$,
2) a virtual gravitational potential gradient to render  gravity compensation $(\nabla V_{\mrm{com,\bar{g}}} =  
mg )$\footnote{$\nabla V_{[.]}$ denotes the gradient of a potential function $V_{[.]}$. For more information 
regarding the Cartesian stiffness and gravitational potentials, see
\cite{Henze2016}.},
3) a feedforward term to improve tracking ($W_\mrm{ff} = M_{\mrm{com}}\Delta \dot{v}_\mrm{com}$) and a compensation term for external 
disturbances  
$-W_\mrm{ext}$ 
\cite{Focchi2018star}:
\begin{subequations}
	\begin{eqnarray}
	W_{\mrm{com,d}} &=& 	W_\mrm{imp} + \nabla V_{\mrm{com,\bar{g}}} + W_\mrm{ff}   - W_\mrm{ext},
	\label{desiredCoMwrench}\\
	W_\mrm{imp} &=& \nabla V_{\mrm{com,K}}   + D_{\mrm{com}} \Delta v_{\mrm{com}},  
	\end{eqnarray}
\end{subequations}
such that $\Delta x_{\mrm{com}} = x_{\mrm{com,d}} - x_{\mrm{com}} $, $\Delta v_{\mrm{com}} = v_{\mrm{com,d}} - 
v_{\mrm{com}}$ and $\Delta \dot{v}_\mrm{com} = \dot{v}_{\mrm{com,d}} - \dot{v}_{\mrm{com}}$ are the tracking 
errors  $\in \Rnum^6$ of the position, velocity and acceleration of the trunk, respectively.

Similarly, the tracking of the swing task is obtained by the virtual force $F_{\mrm{sw,d}} \in \Rnum^{n_{\mrm{sw}}}$.
This is generated by 
\textcolor{black}{1) a Cartesian impedance at the swing foot that is represented by
a stiffness term  ($\nabla V_{\mrm{sw}} = K_{\mrm{sw}}\Delta x_{\mrm{sw}}$)  with positive definite stiffness 
matrix  $K_{\mrm{sw}} \in R^{n_{\mrm{sw}} \times n_{\mrm{sw}}}$
and a damping term ($D_{\mrm{sw}} \Delta v_{\mrm{sw}}$) with positive definite damping matrix $D_{\mrm{sw}} \in 
R^{n_{\mrm{sw}} \times n_{\mrm{sw}}}$, and 2) a feedforward term to improve tracking ($F_{\mrm{sw,ff}}$):}
\begin{eqnarray}
F_{\mrm{sw,d}} &=&  \nabla V_{\mrm{sw}} + D_{\mrm{sw}} \Delta v_{\mrm{sw}} + F_{\mrm{sw,ff}} ,
\label{pass8n}
\end{eqnarray}
such that \textcolor{black}{$\Delta x_{\mrm{sw}} = x_{\mrm{sw,d}} - x_{\mrm{sw}} $ and $\Delta v_{\mrm{sw}} = 
v_{\mrm{sw,d}} - 
v_{\mrm{sw}}$ \textcolor{black}{are} the tracking errors of the swing feet positions and velocities 
respectively}. 
%
%

\subsection{Optimization}
\label{sec:optimization}
%
%
To fulfill the motion tasks in \sref{secTrunkTask} and to distribute 
the load on the stance feet, while respecting the mentioned constraints,
we formulate the \gls{qp}: 
\begin{subequations}
	\begin{eqnarray} 
	& \underset{\vc{u}=[\ddot{\vc{q}}^T \hspace{0.2cm} \vc{\grf}^T]^T}{\min} 	\Vert W_\mrm{com} - W_\mrm{com,d}  
	\Vert^2_Q+ 
	\Vert\vc{u}\Vert^2_{R} \label{cost}\\
  &	\hspace{-2cm}  \st \hspace{0.5cm} \quad \vc{Au} = \vc{b} \label{eqCon},\\
	 &\hspace{-0.6cm} \underline{\vc{d}} < \vc{Cu} < \bar{\vc{d}} \label{ineqCon},
\end{eqnarray}
\label{eq:qp_problem}
\end{subequations}
such that our decision variables  $\vc{u}=[\ddot{\vc{q}}^T \  \vc{\grf}^T]^T \in\Rnum^{6+n+n_{\mrm{st}}}$ 
are the generalized 
accelerations $\ddot{\vc{q}}$ and the contact forces 
$\grf$. 
The cost function \eref{cost} is designed to minimize the \textit{trunk task} and to regularize the solution.
The equality constraints \eref{eqCon} 
encode dynamic consistency, stance constraints and swing tasks. 
The inequality constraints \eref{ineqCon} 
encode
friction constraints, joint kinematic and torque limits. 
All constraints 
are stacked
in the matrix $\vc{A}^T=\mat{\vc{A}_{p}^T & \vc{A}_{\mrm{st}}^T & \vc{A}_{\mrm{sw}}^T}$ and
$\vc{C}^T=\mat{\vc{C}_{\mrm{fr}}^T & \vc{C}_j^T &\vc{C}^T_{\tau}}$ and detailed in
the following sections.
	
\subsubsection{Cost}
The first term of the cost in \eref{cost}
represents the \textit{tracking} error 
between the actual $W_\mrm{com}$  and  the desired $W_\mrm{com,d}$ \gls{com} wrenches from 
\eref{eqn4} and \eref{desiredCoMwrench} respectively.
Since  $W_\mrm{com}$ 
is not a decision
variable, we compute it from the contact forces (see \eref{eqn4}) and
re-write $\|W_{\mrm{com}}-W_{\mrm{com,d}}\|_{\vc{Q}}^2$
in the form of $\|\vc{Gu}-\vc{g}_0\|_{\vc{Q}}^2$ with:
%
	\begin{eqnarray}
G  = \mat{0_{6 \times (6+n)} & J^T_{\mrm{st,com}}}, \quad g_0 =  W_{\mrm{com,d}}.
\end{eqnarray}
\subsubsection{Physical consistency}
\label{sec:physical_contraints}
To enforce physical  consistency between $\grf$ and $\ddot{q}$, we impose the dynamics of the unactuated part of the 
robot
(the trunk dynamics in \eref{eqn4}) as an equality constraint:
\begin{equation}
\vc{A}_p = \mat{\vc{M}_{\mrm{com}}  &0_{6 \times n} & -J^T_{\mrm{st,com}}},\quad
\vc{b}_p = - h_{com}.
\label{eq:physicalEq}
\end{equation}
%
%
%
%

\subsubsection{Stance condition}
\label{sec:stance_contraints}
We can encode the stance feet constraints by re-writing them at the acceleration level in order to be compatible with 
the decision variables. Since $v_{\mrm{st}} = J_{\mrm{st}} \dot{q}$, differentiating once in time, yields to
$\dot{v}_{\mrm{st}} = \vc{J}_{\mrm{st}}\ddot{\vc{q}}+\dot{\vc{J}}_{\mrm{st}}\dot{\vc{q}}$. 
Recalling \assref{ass1} yields $ \vc{J}_{\mrm{st}}\ddot{\vc{q}}+\dot{\vc{J}}_{\mrm{st}}\dot{\vc{q}} = 0$ which is 
encoded as:
%
\begin{equation}
\vc{A}_{\mrm{st}} = \mat{\vc{J}_{\mrm{st}}  & \vc{0}_{n_{\mrm{st}} \times n_{\mrm{st}} }}, \quad
\vc{b}_{\mrm{st}} = -\dot{\vc{J}}_{\mrm{st}}\dot{\vc{q}},
\label{eq:stanceEq}
\end{equation}
such that $\dot{\vc{J}}_{\mrm{st}}$ is the time derivative of $J_{\mrm{st}}$.
For numerical precision, we
compute the product $\dot{\vc{J}}_{\mrm{st}}\dot{\vc{q}}$ using spatial algebra.
\subsubsection{Swing task}
\label{sec:swingTask}
Similar to \sref{sec:stance_contraints}, we can encode the swing task directly as an
\textit{equality constraint}, i.e. by enforcing the swing feet to
follow a desired swing acceleration
$\dot{\vc{v}}_{\mrm{sw}}(\vc{q})=\dot{v}_{\mrm{sw,d}}\in\Rnum^{n_{\mrm{sw}}}$ 
yielding:
\begin{equation}
\vc{J}_{\mrm{sw}}\ddot{\vc{q}} + \dot{\vc{J}}_{\mrm{sw}}\dot{\vc{q}} =
\dot{\vc{v}}_{{\mrm{sw,d}}},
\end{equation}
that in matrix form becomes\footnote{Alternatively, it is possible to write the
swing task at the joint space rather than in the operational space by changing
the matrix $\vc{A}_{\mrm{sw}}, \vc{b}_{\mrm{sw}}.$}:
\begin{equation}
\vc{A}_{\mrm{sw}} = \mat{\vc{J}_{\mrm{sw}}  & \vc{0}_{n_{\mrm{sw}}\times n_{\mrm{st}}}}, \quad
\vc{b}_{\mrm{sw}} = \dot{\vc{v}}_{{\mrm{sw,d}}} -\dot{\vc{J}}_{\mrm{sw}}\dot{\vc{q}}.
\label{eq:swingEq}
\end{equation}
%
%
%
%
Note that this implementation is analogous to the \textcolor{black}{swing} task in \sref{secTrunkTask}. 
The difference is that this implementation is at the 
acceleration level while the other is at the force level. 
%
%
In Section \ref{sec:slacks}
we incorporate slacks in the optimization to allow temporary violation of the swing
tasks (e.g. useful when the kinematic limits are reached).
\subsubsection{Friction cone constraints}
\label{sec:friction_constraints}
To avoid slippage and obtain a smooth
loading/unloading of the legs, we incorporate friction constraints. For that, we ensure that the contact forces lie
inside the friction cones and their normal components stay within some
user-defined values (i.e. maximum and minimum force magnitudes). We approximate
the friction cones with square pyramids to express them with linear constraints.
The fact that the ground contacts are unilateral, 
can be naturally encoded by setting an  ``almost-zero''
lower bound on the normal component, 
while the upper bound allows us to regulate
the amount of ``loading'' for each leg. We define the friction inequality constraints as:
\begin{align}
\underline{\vc{d}}_{\mrm{fr}} < \vc{C}_{\mrm{fr}}\vc{u} < \bar{\vc{d}}_{\mrm{fr}}, \quad
\vc{C}_{\mrm{fr}} =
\mat{\vc{0}_{p\times(6+n)} & \vc{F}_{\mrm{fr}}},
\label{eq:frictionIneq}
\end{align}
with:
\begin{equation}
\vc{F}_{\mrm{fr}} = \mat{\vc{F}_0 & \dots & \vc{0} \\
	\vdots & \ddots & \vdots \\
	\vc{0} & \dots & \vc{F}_c},  \ \ \
\underline{\vc{d}}_{\mrm{fr}} = \mat{\underline{\vc{f}}_0 \\
	\vdots \\
	\underline{\vc{f}}_c}, \ \ \
\bar{\vc{d}}_{\mrm{fr}} = \mat{\bar{\vc{f}}_0 \\
	\vdots \\
	\bar{\vc{f}}_c},
\end{equation}
where $\vc{F}_{\mrm{fr}}\in\Rnum^{p\times n_\mrm{st}}$ is a block diagonal matrix encoding the friction cone
boundaries for each stance leg and
$\underline{\vc{d}}_{\mrm{fr}},\bar{\vc{d}}_{\mrm{fr}}\in\Rnum^{p}$ are the lower/upper
bounds respectively. For the detailed  implementation of the friction constraints refer to \cite{focchi2017auro}.


\subsubsection{Torque limits}
\label{sec:torque_constraints}
We encode the torques from the decision
variables since they can be expressed as a bi-linear function
of $\ddot{q}_j$ and $\grf$. 
For that, we constrain the joint torques (the actuation limits $\vc{\tau}_{\min} < \vc{\tau}_j <
\vc{\tau}_{\max}$) by exploiting the actuated part of the full dynamics
\eref{eq:full_dynamicsCOM}:
\begin{subequations}
	\begin{eqnarray}
\underline{\vc{d}}_{\tau} < \vc{C}_{\tau}\vc{u} < \bar{\vc{d}}_{\tau},\quad
 \vc{C}_{\tau} = \mat{0_{n \times 6} & \vc{\bar{M}}_j & - \vc{J}_{\mrm{st,j}}^T},\\
\underline{\vc{d}}_{\vc{\tau}} = - \bar{\vc{h}}_j + \vc{\tau}_{\min}(\vc{q}_j), \quad 
\bar{\vc{d}}_{\vc{\tau}} = - \bar{\vc{h}}_j +  \vc{\tau}_{\max}(\vc{q}_j),
\label{eq:frictionIneq}
\end{eqnarray}
\end{subequations}
where $\vc{\tau}_{\min}(\vc{q}_j), \vc{\tau}_{\max}(\vc{q}_j)\in\Rnum^n$ are
the lower/upper bounds on the torques. In the case of our quadruped robot, 
these bounds must be recomputed at each
control loop because they  depend on the joint positions. This is due to
the presence of linkages on the sagittal joints (\textcolor{black}{HFE and KFE}), 
that set a joint\textcolor{black}{-}dependent profile on the maximum torque
 (non-linear in the joint range).

\subsubsection{Joint kinematic limits}
\label{sec:kinematic_constraints}
We enforce joint kinematic constraints as function of the joint accelerations
(i.e. $\ddot{\vc{q}}_{j_{\min}} < \ddot{\vc{q}}_j < \ddot{\vc{q}}_{j_{\max}}$).
We select them via the matrix $\vc{C}_j$:
\begin{subequations}
	\begin{eqnarray}
\underline{\vc{d}}_j < \vc{C}_j \vc{u} < \bar{\vc{d}}_j,\quad
\vc{C}_j = \mat{\vc{0}_{n\times 6}  &  \vc{I}_{n\times n}  &  \vc{0}_{n
		\times n_{\mrm{st}}}},\\
 \underline{\vc{d}}_j =  \ddot{\vc{q}}_{j_{\min}}(\vc{q}_j), \quad
\quad \bar{\vc{d}}_j = \ddot{\vc{q}}_{j_{\max}}(\vc{q}_j),
\label{eq:jointKinIneq} 
\end{eqnarray} 
\end{subequations}
such that $\ddot{\vc{q}}_{j_{\min}}(\vc{q}_j)$  and $\ddot{\vc{q}}_{j_{\max}}(\vc{q}_j)$ are the upper/lower bounds on 
accelerations. These bounds 
should be recomputed at each control
loop. They are set  in order to make the joint reach the end-stop at a zero
velocity in a time interval $\Delta t = 10dt$, where $dt$ is the loop
duration. For instance, if the joint is at a distance $ 
\vc{q}_{j_{\max}}-\vc{q}_j$ from the end-stop with a velocity $\dot{\vc{q}}_j$, the
deceleration to cover this distance in a time interval  $\Delta t$, and approach
the end-stop with zero velocity, will be:
\begin{equation}
\ddot{\vc{q}}_{j_{\min,\max}} = -\frac{2}{\Delta t^2}(\vc{q}_{j_{\min,\max}} -
\vc{q}_j - \Delta t\, \dot{\vc{q}}_j).
\label{eq:accellBounds}
\end{equation}

\subsection{Torque computation}
\label{secTorqueComp}
We map the optimal solution  $\vc{u}^* = \mat{\ddot{\vc{q}}^* & 
\vc{\grf}^*}$ obtained by solving \eref{eq:qp_problem}, 
into desired  joint torques
$\vc{\tau}_{d}^{*} \in\mathbb{R}^n$ using the actuated part of the full
dynamics equation of the robot as:
\begin{equation}
\vc{\tau}_{d}^{*} = \bar{M}_j\ddot{\vc{q}_j}^* + \bar{\vc{h}}_j -
\vc{J}_{\mrm{st,j}}^T\vc{\grf}^*
\label{eq:torques}.
\end{equation}
\section{Passivity Analysis}  
\label{sec:Passivity}    
The overall system consists of the \gls{wbc}, the robot and 
the environment.
This system is said to be \textit{passive} if all these components, and their interconnections are passive 
\cite{Stramigioli2015}.
If the robot and the environment are passive, and the controller is proven to be passive, then the overall 
system is passive \cite{Schaft}. 
A system (with input $u$ and output $y$) is said to be passive if there exists a storage function $S$ that is bounded 
from below and its derivative $\dot{S}$ is less than or equal to its supply rate ($s=y^Tu$). 
In this context, we define the total energy stored in the controller to be the candidate storage function for the 
controller $S=V$.
The rest of this section is devoted to analyze the passivity of the overall system. 
%
\begin{assumpB}
	 A feasible solution exists for the \gls{qp} in \eref{eq:qp_problem} 
	 in which the motion 
	 tasks are achieved. Moreover, we do not consider the feed-forward terms in \eref{desiredCoMwrench} and \eref{pass8n}  leaving 
	 this to future developments.   
	 \label{ass2}
\end{assumpB} 

We start by defining the velocity error at the joints and at the stance feet to be $\Delta \dot{q}_j= 
\dot{q}_{j,d} - \dot{q}_\nmrm{j}$, and $\Delta v_{\mrm{st}} = v_{\mrm{st,d}}-v_{\mrm{st}}$, respectively. 
We also define the  desired feet forces $F_\mrm{d}= [F_{\mrm{st,d}}^T \ F_{\mrm{sw,d}}^T]^T$ such that, 
by following the sign convention in \fref{fig:conventions}, 
the mapping between $F_\mrm{d}$ and $W_\mrm{com,d}$
is expressed as%
\footnote{\textcolor{black}{Since we are analyzing the passivity of the controller,
we are interested in the forces exerted by the robot on the environment rather
than the forces exerted by the environment on the robot. Hence the mapping in \eref{sign} is  
negative}.}:
\begin{equation}
W_\mrm{com,d} =  - J_\mrm{com}^T F_\mrm{d} \label{sign},  \end{equation}
while mapping between $F_d$ and $\tau_j$   is expressed as\footnote{Assuming a perfect low level torque control 
tracking (i.e., $\tau_\mrm{d} = \tau$).}
\begin{eqnarray}
\tau &=&  J_{\nmrm{j}}^T  F_d \label{torquemapp} .\label{pass6}
\end{eqnarray}
%
%
By defining $\nabla V_\mrm{com} = \nabla V_\mrm{com,K} + \nabla V_\mrm{com,\bar{g}}$ and recalling \eref{sign}, we 
rewrite  
\eref{desiredCoMwrench} under  \assref{ass2} as: 
\begin{subequations}
	\begin{eqnarray}
	\nabla V_{\mrm{com}} &=& 	W_{\mrm{com,d}}  - D_{\mrm{com}} \Delta v_{\mrm{com}} \label{pass9b}\\
	&=& - J_{\mrm{com}}^{T} F_\mrm{d} - D_{\mrm{com}}  \Delta v_{\mrm{com}} \label{pass9c}.
	\end{eqnarray}
\end{subequations}
\subsection{Analysis}
The overall energy in the whole-body controller is the one stored in the virtual impedance 
at the CoM and the potential energy due to gravity compensation ($V_{\mrm{com}}$),
and the energy stored in the virtual impedances at the swing feet ($ V_{\mrm{sw}}$): 
\begin{equation}
V =  V_{\mrm{com}} + V_{\mrm{sw}} .
\end{equation}
The time derivatives are\footnote{The time derivative of an arbitrary storage function $\dot{V}(\Delta x(t))$ that is a 
function of $\Delta x(t)$ could be written as $\dot{V} = \frac{d}{dt} \Delta x^T(t) \cdot \frac{\partial}{ \partial 
\Delta x(t)} V$ that is written for short as $\dot{V} = \Delta v^T \nabla V $.}:
\begin{equation}
\dot{V} = \dot{V}_{\mrm{com}} + \dot{V}_{\mrm{sw}} = \Delta v_\mrm{com}^T  \nabla V_{\mrm{com}}  + \Delta 
v_\mrm{sw}^T \nabla V_{\mrm{sw}}
\label{vdot}.
\end{equation}
Recalling \eref{pass8n} and \eref{pass9c}, \eref{vdot} yields:
\begin{eqnarray}
\dot{V} &=& \Delta v_\mrm{com}^T (-J_{\mrm{com}}^{T} F_\mrm{d} - D_\mrm{com} \Delta v_\mrm{com}) + 
\nonumber\\
&~& \Delta v_\mrm{sw}^T (F_\mrm{sw,d} - D_\mrm{sw} \Delta v_\mrm{sw}) ) .
\end{eqnarray}
We regroup $\dot{V}$ in terms of the non-damping terms $\dot{V}_1$ and damping terms  $\dot{V}_2$ yielding:
\begin{subequations}
	\begin{eqnarray}
	\dot{V_1} &=& - \Delta v_\mrm{com}^TJ_{\mrm{com}}^{T} F_\mrm{d}  + \Delta v_\mrm{sw}^T 
	F_\mrm{sw,d} \label{pass12aa} \\
	\dot{V_2} &=& - \Delta v_\mrm{com}^T D_\mrm{com} \Delta v_\mrm{com} - \Delta v_\mrm{sw}^T 
	D_\mrm{sw} \Delta v_\mrm{sw} \label{pass12aaa}.
	\end{eqnarray}
\end{subequations}
We rewrite \eref{eq_pass4} in terms of $\Delta v_\mrm{com}$, $\Delta v$ and $\Delta \dot{q}_\mrm{j}$ as:
\begin{subequations}
	\begin{eqnarray}
	\Delta	v &=&  J_{\mrm{com}} \Delta v_{\mrm{com}} +   J_{\mrm{j}} \Delta \dot{q}_\mrm{j}  \\
	\Delta v_{\mrm{com}}^T  J_{\mrm{com}}^T &=& - \Delta \dot{q}_\mrm{j}^T J_{\mrm{j}}^T + \Delta v^T.
	\label{pass14c}
	\end{eqnarray}
\end{subequations}
Plugging, \eref{pass14c} in \eref{pass12aa} yields\footnote{From the definition of  $v$ and $F_d$, we get $v^TF_d = 
v_{\mrm{st}}^T F_{\mrm{st,d}} +  v_{\mrm{sw}}^T 
	F_{\mrm{sw,d}} $.}:
\begin{subequations}
	\begin{eqnarray}
\dot{V_1} &=&  \Delta \dot{q}_\mrm{j}^T J_{\mrm{j}}^T F_\mrm{d} - \Delta v^T F_\mrm{d}  + \Delta 
v_\mrm{sw}^T F_\mrm{sw,d} \\
&=& \Delta \dot{q}_\mrm{j}^T J_{\mrm{j}}^T F_\mrm{d}   - \Delta v_\mrm{st}^T F_\mrm{st,d} \label{pass15a} .
\end{eqnarray}
\end{subequations}
Plugging \eref{pass6} and into \eref{pass15a} yields%
:
\begin{eqnarray}
\dot{V_1} &=&  \Delta \dot{q}_\mrm{j}^T \tau  - \Delta v_\mrm{st}^T F_\mrm{st,d} \label{pass16a} .
\end{eqnarray}
Under  \assref{ass1}, \eref{pass16a} yields:
\begin{eqnarray}
\dot{V_1} &=&   \Delta \dot{q}_\mrm{j}^T \tau \label{pass17a} .
\end{eqnarray}
Thus, $\dot{V}$ could be rewritten as:
\begin{eqnarray}
\dot{V} =  \Delta \dot{q}_\mrm{j}^T \tau - \Delta v_{\mrm{com}}^T D_{\mrm{com}}\Delta v_{\mrm{com}}    - \Delta 
v_{\mrm{sw}} ^T D_{\mrm{sw}}\Delta v_{\mrm{sw}}.
\label{pass92}
\end{eqnarray}
\vspace{-0.75cm}
\subsection{Proof}
Under \assref{ass2}, the designed \gls{wbc} is an impedance control with 
gravity compensation 
that, similar to a PD+ \cite{Ortega2013}, defines a map of $( \dot{q}_\mrm{j} -
\dot{q}_\mrm{j,d}) \mapsto - \tau$\footnote{Note that  $ \dot{q}_\mrm{j} - \dot{q}_\mrm{j,d}  = -\Delta \dot{q}_\mrm{j}$.
	Thus, the controller with the map $( \dot{q}_\mrm{j} - \dot{q}_\mrm{j,d}) \mapsto - \tau $ 
	has a supply rate of $ \Delta \dot{q}_\mrm{j}^T \tau$.}. 
This controller is passive if $V$ is bounded from below and $\dot{V}\leq   \Delta \dot{q}_\mrm{j}^T \tau$. 
Since $V$ consists of positive definite potentials that resemble Cartesian 
stiffnesses at the CoM and the swing leg(s), and under the assumption
 the gravitational potential is bounded from below (see \cite{Ott2008}), 
 $V$ is also bounded from below. Additionally, recalling \eref{pass92} 
 proves that the controller is indeed passive; thus, the overall system is passive. 

\section{Implementation Details}
\label{sec:Implementations}
This section  describes some pragmatic details 
that we found crucial in the implementation on the real platform. 
\vspace{-0.5cm}
\subsection{Stance task}
\label{sec:stanceAccel}
%
Uncertainties in estimating the terrain's normal
direction and friction coefficient could result in slippage. This can lead 
to considerable  motion of the stance feet with  an eventual loss of stability.
%
%
To avoid this, a joint impedance feedback loop could be run in parallel to the \gls{wbc},
at the price of losing the capability of optimizing the \gls{grfs}.
%
%
%
A cleaner solution is to incorporate in the optimization  Cartesian impedances, specifically designed
to keep the relative distance among the stance feet constant (we denote it \textit{stance task}).

\textit{The stance task} \textcolor{black}{has an influence} \textit{only} when there is 
an anomalous motion in the stance feet, retaining
the possibility to freely optimize for \gls{grfs} in normal situations.
This can be achieved by re-formulating the stance
condition in \eref{eq:stanceEq} \textcolor{black}{as a} desired 
stance feet acceleration   $\dot{v}_{\mrm{st,d}}$ as:
\begin{equation}
\dot{v}_{\mrm{st,d}}  =   K_{\mrm{st}}(\hat{x}_{\mrm{st}} - x_{\mrm{st}}) - D_\mrm{st}v_{\mrm{st}}
\label{eq:stanceAccelerations}.
\end{equation} 

\textcolor{black}{This term} is added to $b_{st}$ (in \eref{eq:stanceEq}) as $b_{st} = -\dot{J}_{st}\dot{q} + 
\ddot{x}^d_{f_{st}} $ 
%
%
where  $\hat{x}_{\mrm{st}}$ is a sample of the foot 
position at the touchdown (in the world frame).
%
\vspace{-0.5cm}
\subsection{Constraint Softening}
\label{sec:slacks}
Adding  slack variables to an optimization problem
is commonly done to avoid infeasible solutions,
by allowing a certain degree of constraint violation. 
Infeasibility can occur when hard constraints are conflicting with each other,
which can be the case in our \gls{qp}. 
%
%
%
Hence, some of the equalities/inequalities in \eref{eq:qp_problem}
 should be relaxed if they are in conflict.

We decided not to introduce slacks in the torque constraints (\sref{sec:torque_constraints})
or the dynamics   (\sref{sec:physical_contraints}) keeping them as hard 
constraints, since torque constraints and physical consistency should never be violated. 
%
On the other hand  it is important to allow a certain level
of relaxation for the swing tasks in \sref{sec:swingTask} 
that could be violated if the joint kinematic limits are reached%
%
\footnote{Using slacks in friction constraints did not result in  significant improvements. 
	}.

To relax the constraints of the swing task, 
first, we augmented the decision variables 
$\vc{\tilde{u}} = [\vc{u}^T \ \epsilon^T]^T \in \Rnum^{6+n+k+n_{sw}}$  
with the vector of slack variables  
$\epsilon \in \Rnum^{n_{sw}}$ where we introduce a slack variable  for each direction of the swing tasks. 
Then, we replaced the swing task  equality constraint 
$\vc{A}_{sw} u + \vc{b}_{sw}=0$,  by two inequality 
constraints: 
%
\begin{align}
\label{eq:slackIneq}\nonumber
-\epsilon \leq \vc{A}_{sw} \tilde{u} &+ \vc{b}_{sw} \leq \epsilon\\
\epsilon & \geq 0.
\end{align}
%
The first inequality in \eref{eq:slackIneq} restricts  the solution to a bounded region around  the original 
constraint while the second one ensures that the slack variables remain non-negative. 
%
%
%
To make sure that there is  a constraint violation only when  the constraints are
conflicting, we minimize the norm of the slack vector adding a 
regularization term $\alpha\Vert \epsilon \Vert^2$ to \eqref{cost}
with a high weight $\alpha$.
 
To reduce computational \textcolor{black}{complexity}, 
we could have introduced a single slack for each swing task (instead of 
one for each direction).
However, this could create coupling errors in the tracking. 
For instance, since the \gls{hfe} joint (see \fref{fig:conventions}) mainly \textcolor{black}{acts} in the $XZ$ plane, 
if it reaches its joint limit, only that plane should be affected leaving the $Y$ direction unaffected.
%
A single slack couples the three directions causing tracking errors also in the $Y$ direction.
%
Conversely, using multiple slacks,  only the directions  the \gls{hfe} acts upon, will be affected.

\vspace{-0.25cm}
\section{Results}
\label{sec:expResults}
In this section we validate the capabilities of the
controller under various terrain conditions and locomotion gaits.
The \gls{wbc} and torque control loops run
in real-time threads at \unit[250]{Hz} and \unit[1]{kHz}, respectively.
We set the gains for the swing tasks 
to $\vc{K}_{sw} = \mrm{diag}({2000,2000,2000})$ and $\vc{D}_{sw} =
\mrm{diag}({20,20,20})$,
while for the trunk task we set $\vc{K}_{x} = \mrm{diag}({2000,2000,2000})$
$\vc{D}_{x} = \mrm{diag}({400,400,400})$ and $\vc{K}_{\theta} = \mrm{diag}({1000,1000,1000})$
$\vc{D}_{\theta} = \mrm{diag}({200,200,200})$. These values proved to be working
in both: simulation and real experiments. The results are collected in the accompanying 
video\footnote{\href{https://www.dropbox.com/s/pukm88fd3spjqhi/passive_wbc.mp4}{\text{https://www.dropbox.com/s/pukm88fd3spjqhi/passive\_wbc.mp4}}}.
Additionally, in experimental trials, we also included a low gain joint-space attractor
(PD controller) for the swing task since \textcolor{black}{imprecise} torque tracking of the 
knee joints (due to the low inertia) produce control instabilities in an
operational space implementation (e.g. the one in Section \ref{sec:swingTask}).
%
%

\subsection{Constraint Softening through Slack Variables}
\begin{figure}[tb]
	\centering
	\includegraphics[width=\columnwidth]{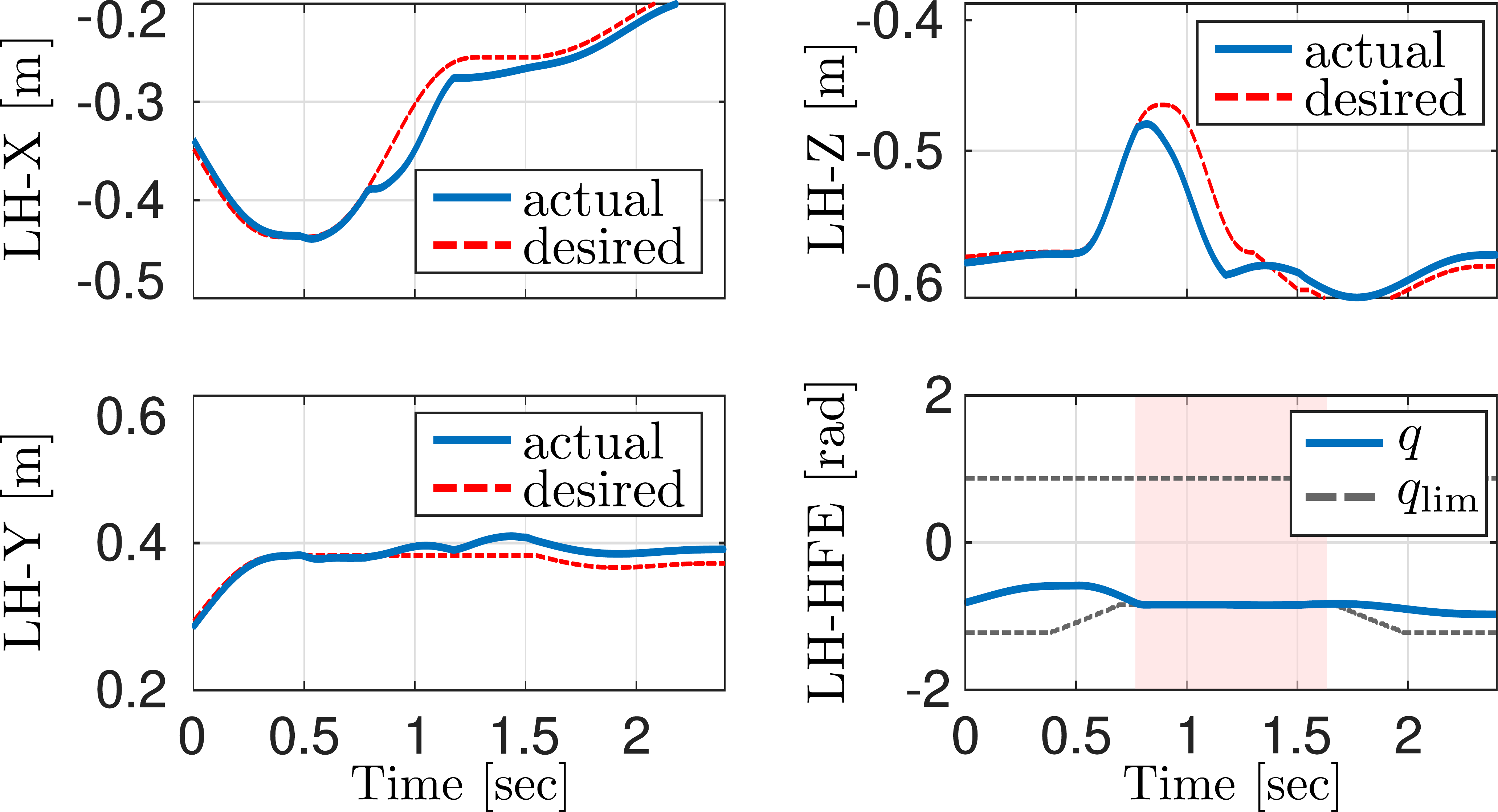}
	\reducespace
	\caption{ \small Simulation. Effect of (kinematic limits) slacks variables on foot tracking. The
	upper-left/right and bottom-left plots show the tracking of the desired foot
	position (\acrshort{lh} leg) in $X$, $Y$ and $Z$, respectively. Bottom-right
	plot depicts the joint limits (black line) and the actual position (blue line) of the \gls{hfe} joint.
	The red shaped area underlines that the \gls{hfe} joint is properly clamped when the slack increases. \vspace{-0.5cm}}
	\label{fig:effectOfSlacks}
\end{figure} 
In \fref{fig:effectOfSlacks} we artificially incremented the lower limit of the
\acrshort{lh}-\gls{hfe} joint.
When the limit is hit, the bound on the joint acceleration
\eref{eq:accellBounds} produces a desired torque command that stops its motion.
This ``naturally'' clamps the actual joint position to the limit (bottom-left
plot) and   influences the  foot tracking mainly  along the $X$ and $Z$ directions.
%
%

\textit{Computational time:} the solution of the \gls{qp}
takes between \unit[90-110]{$\mu s$}  on a Intel  i5 machine without the
slacks variables.
After augmenting the problem with the slack variables and its constraints, 
it increases   \unit[30]{\%} on average (\unit[120-150]{$\mu s$}). However it still
remains suitable for real-time implementation (250 $Hz$).
\subsection{Friction Constraints and Bounded Slippage}
\label{sec:frictionTests}
We evaluated in simulation the controller performance against inaccurate 
friction coefficient estimates $\mu$, which define incorrectly the friction cone
constraints in the \gls{wbc}.
In the accompanying video, we show an example where the robot crawls at
\unit[0.11]{m/s} on a slippery floor ($\mu=0.4$) while we set the friction
coefficient to $\mu=1.0$ in the \gls{wbc}, to emulate an estimation error. 
Simulation results support the fact that foot slippage remains bounded
by the action of the \textit{stance task} (\sref{sec:stanceAccel}).
If we gradually correct $\mu$ the slippage events
completely disappear; allowing an increase of forward velocity up to \unit[0.16]{m/s}.
%
%
%
\subsection{Torque Limits and Load Redistribution}
\label{sec:expTorqueLim}
\begin{figure}[tb]
	\centering
	\includegraphics[width=\columnwidth]{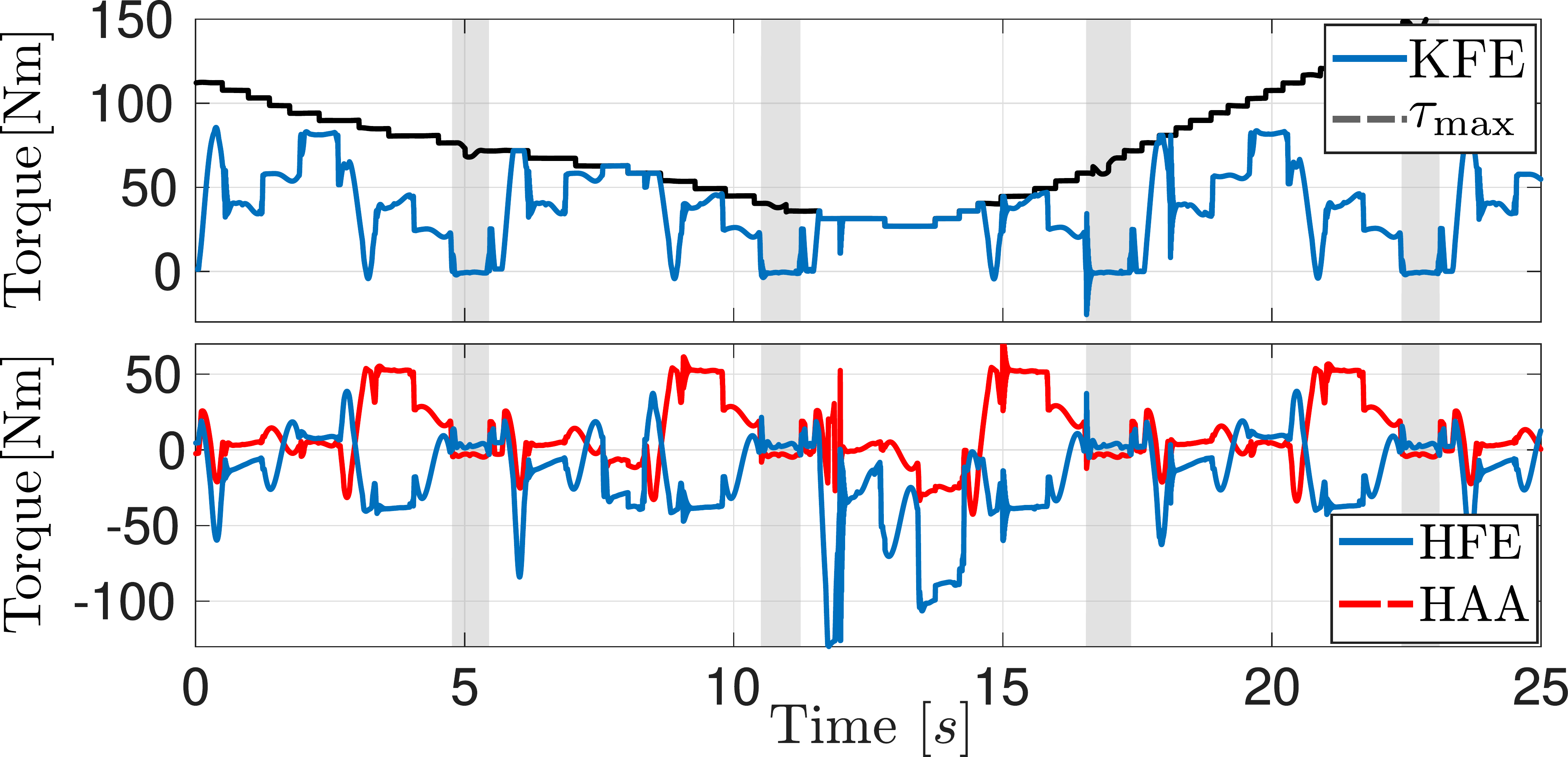}
	\reducespace
	\caption{\small Simulation. Effect of  introducing an artificial torque limit on the \acrshort{kfe} 
	joint \textcolor{black}{of the LF leg} during a typical crawl. 
	\textcolor{black}{The shaded area represents the swing phase of the leg while the unshaded part is the stance 
	phase. }
	 We reduced the torque limit down to \unit[26]{Nm}
	(black line, upper plot), and as a consequence the
	\acrshort{hfe} torque is increased by the controller (bottom plot).}
	\label{fig:torqueLimits}
\end{figure}

We analyzed in simulation the effect of adding  an artificial torque limit,  in our \gls{wbc}. 
This helps us to derive controllers that are robust against joint damages.
Figure \ref{fig:torqueLimits} shows a reduction of the torque limits down to
\unit[26]{Nm} in the \textcolor{black}{\gls{lf}}-\acrshort{kfe} joint and the load redistribution among
the other joints (\acrshort{haa} and \acrshort{hfe}) \textcolor{black}{of the LF leg}.
Indeed, while the \acrshort{kfe} joint torque is clamped, the \gls{hfe} is
loaded more (lower plot). This load redistribution did not affect the trunk 
motion and it demonstrates how the controller exploits the torque redundancy by
finding a new load distribution.
%

\begin{figure}[tb]
	\centering
	\includegraphics[width=\columnwidth]{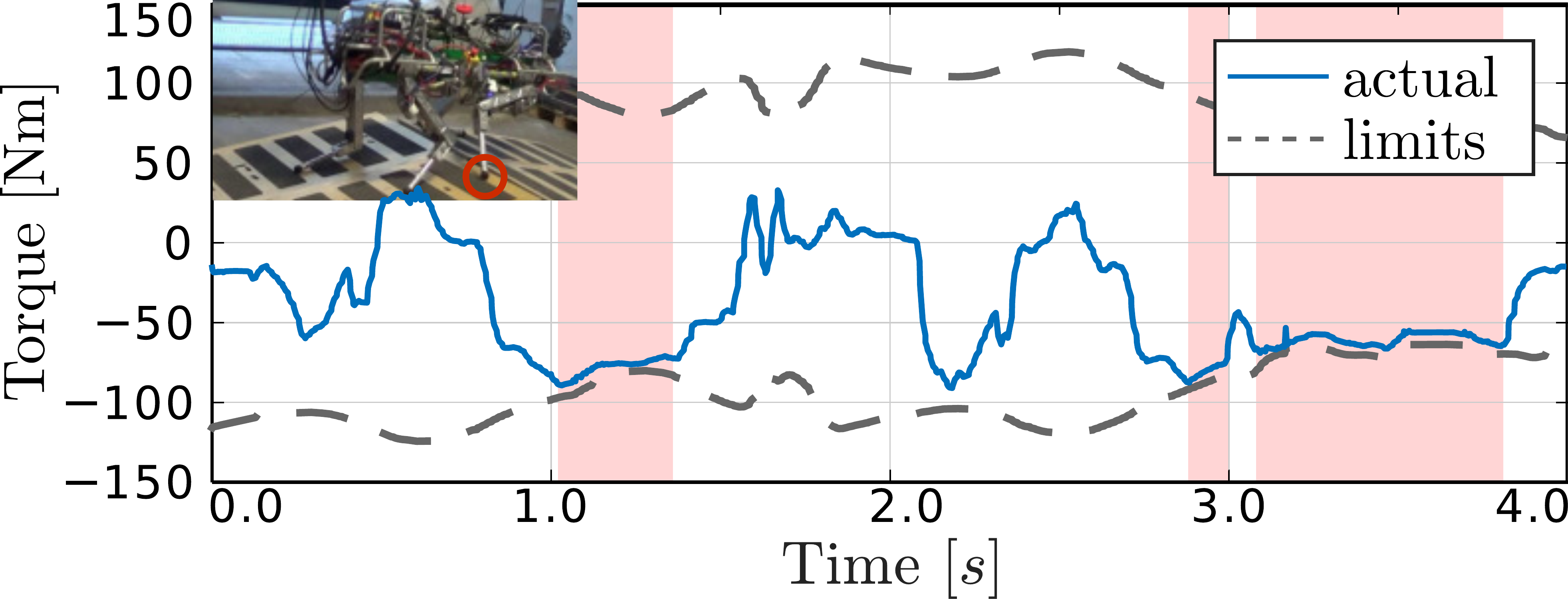}
	\reducespace
	\caption{\small Real experiments. Reaching the torque limits on the RF-HFE  joint
	while climbing up and down two ramps. The \acrshort{hyq} robot reached three times its
	torque limits (red shaded areas). The real torque (blue line) is tracking
	the desired one (not shown) computed from our whole-body controller while satisfying the
	joint limits (black line). The torque limits are time-varying due to the
	joint mechanism.}
	\label{fig:reachingLimits}
\end{figure}

We carried out also intensive experimental validation in various challenging
terrains. Slopes increase the probability of reaching torque limits because 
of the more demanding kinematic configurations. Indeed, in
 \fref{fig:reachingLimits}, the robot reached three times the torque limits (red
shaded areas). Crossing this terrain would not be possible without enforcing
the torque limits as hard constraints.

\subsection{Different Torque Regularization Schemes}
\label{sec:torqueReg}
By setting different regularizations in  \eref{cost} 
\cite{focchi2017auro}, we can either choose to 
maximize the robustness to uncertainties in the friction
parameters (e.g. \gls{grfs} closer to the friction cone normals) or to minimize
the joint torques\footnote{Setting the weighting matrix $R_{kk} = J_{st} S^T
R_{\tau}S J_{st}^T$ where: $R_{kk}$ is the sub-block of $R$  to
the \gls{grfs} variables and $S$ selects the actuation joints.}.
In the latter case, for instance, we could encourage the controller to use a
particular joint by increasing its corresponding weight.
If we gradually increase the weight of the \gls{kfe} joints (see accompanying
video), the effect of torque regularization becomes visible because the
\gls{grfs} are no longer vertical. Indeed the \gls{grfs} start to point toward
the knee \textcolor{black}{axis} in order to reduce its torque command.

%

%
%
\subsection{Comparison with Previous Controller (Quasi-Static)}
\label{sec:staticVSDynamic}
\begin{figure}[!thb]
	\centering
	\includegraphics[width=\columnwidth]{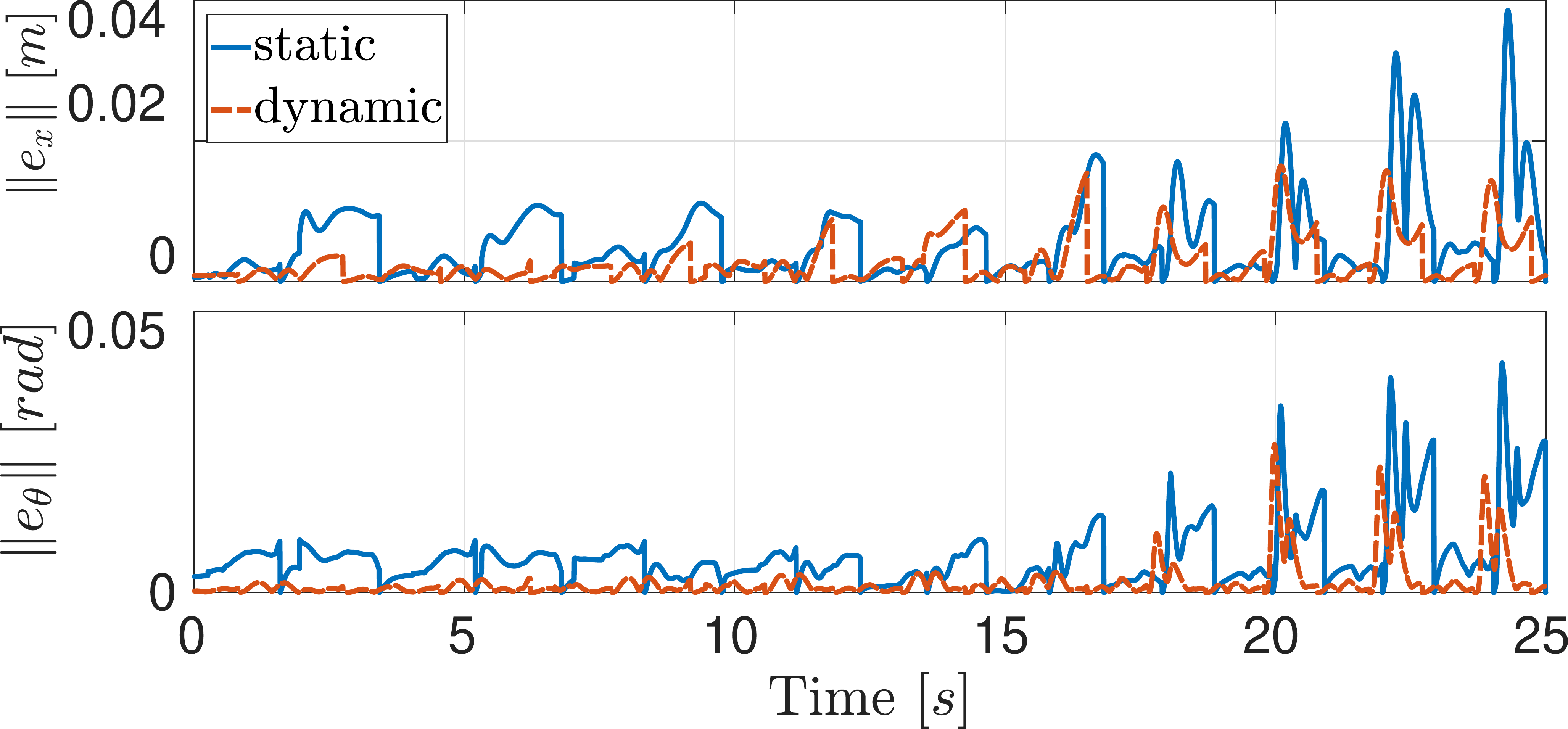}
	\reducespace
	\caption{\small Simulation. Comparison of tracking errors for the trunk task of a
	quasi-static controller \cite{focchi2017auro} against our whole-body controller
	(dynamic). $l^2$-norms of linear and angular errors are shown in the top and bottom figures.
	Note that the errors are reset to zero at each step due to re-planning \cite{Focchi2018star}.}
	\label{fig:staticVSdynamic}
\end{figure}

We compare our whole-body controller (dynamic) against a centroidal-based
controller (quasi-static) \cite{focchi2017auro}.
As metric we use the $l^2$-norm for the linear $ e_{x} $ and angular
$ e_{\theta}$ tracking errors of the trunk task.
If we increase linearly the forward speed from \unit[0.04]{m/s} to
\unit[0.15]{m/s}, the tracking error is reduced approximately by \unit[50]{\%} in
comparison to the quasi-static controller (\fref{fig:staticVSdynamic}).
This is due to the fact that our \gls{wbc} computes both joint accelerations
and contact forces, which allows a proper mapping of torque commands
(inverse dynamics).
Indeed this results in accurate execution of more dynamic motions.
%
%
%
%
\subsection{Disturbance Rejection against Unstable Foothold}
\begin{figure}[tb]
	\centering
	\includegraphics[width=\columnwidth]{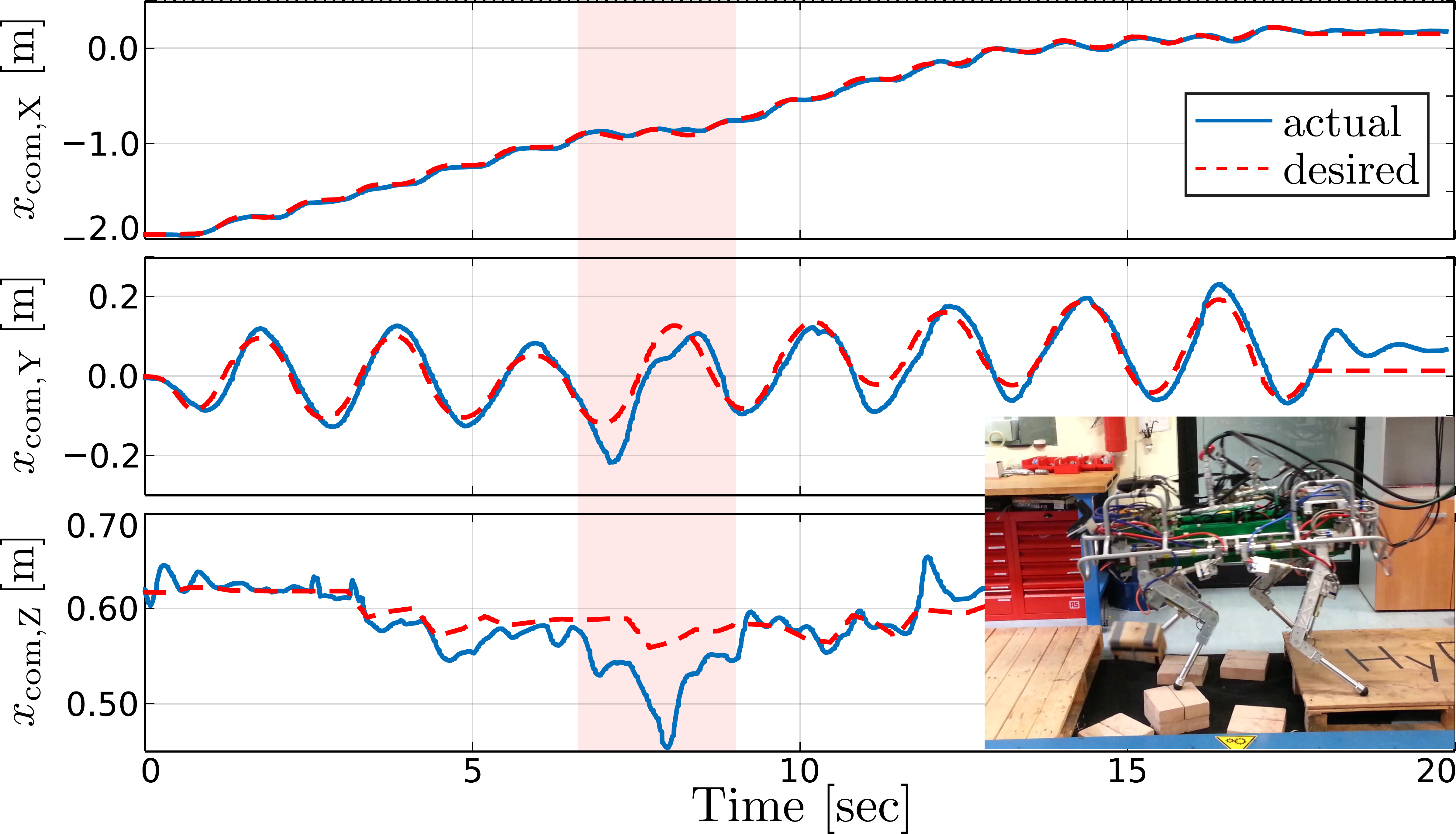}
	\reducespace
	\caption{\small Real experiments. Disturbance rejection against unstable foothold that occurs when a
	stepping-stone rolled \textcolor{black}{under the RF leg} at \unit[$t=6.5$]{s}. The controller lost
	tracking of the \gls{com} height, however, the friction cone constraints keep
	instantaneously the robot's balance. Indeed a good tracking of the horizontal
	motion of the \gls{com} is obtained. Note that the red shaded area depicts
	the moments of the disturbance rejection.}
	\label{fig:complianceRejection}
\end{figure}

We encode compliance tracking of the \gls{com} task through a
\textit{virtual} impedance.
Friction cone constraints help to instantaneously keep the robot's balance
whenever a tracking error happens due to, for instance, an unstable foothold.
Furthermore joint constraints (positions and torques) guarantee feasibility of
the computed torque commands. Figure \ref{fig:complianceRejection} shows how the controller
compliantly tracks the desired \gls{com} trajectory during an unstable footstep (a
rolling stepping-stone) that occurs at \unit[$t=6.5$]{s}. 
This creates  tracking errors on the \gls{com} height, yet, 
good tracking performance is kept for the horizontal \gls{com} motion, due to
friction cone constraints that maintained the robot's balance along the entire
\textcolor{black}{locomotion}. 
%
\vspace{-0.1cm}
\subsection{Locomotion over Slopes and Terrain Mapping}
\begin{figure}[tb]
	\centering
	\includegraphics[width=0.48\textwidth]{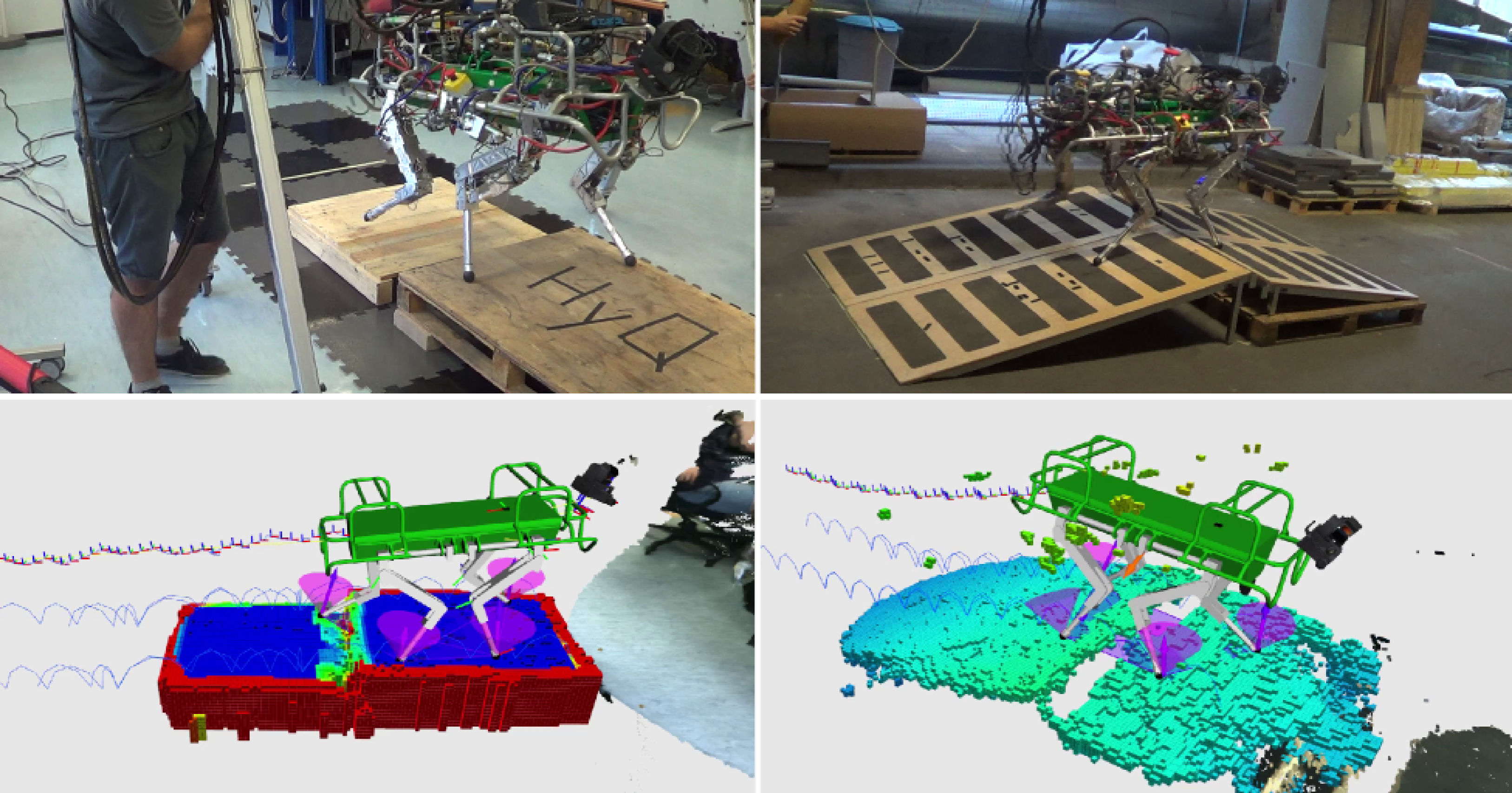}
	\caption{\small Snapshots of experimental trials to evaluate our whole-body control
	and online terrain mapping. Left column: crossing a \unit[22]{cm} gap with a
	\unit[7]{cm} step. Right column: traversing two ramps with a \unit[15]{cm}
	gap between them.}
	\label{fig:controlMapping}
\end{figure}

These experiments have been performed with online terrain mapping
\cite{Focchi2018star}. Both the terrain mapping and the whole-body controller make
use of a drift-free state estimation algorithm to obtain the body state
\cite{nobili2017a}. The friction cone constraints of the controller are
described given the real terrain normals provided by an onboard mapping algorithm\footnote{The controller action can be 
greatly improved by setting the real terrain normal (under each foot)
rather than using a default value for all the feet.}.
The friction coefficient has been conservatively set to 0.7 \textcolor{black}{for all the experiments}. Figure 
\ref{fig:controlMapping} shows different snapshots of
various challenging terrain used to evaluate our controller.
The centroidal trajectory, gait and footholds are computed simultaneously
as described in \cite{Cabezas2018}.

\subsection{Tracking Performance with Different Gaits}
\begin{figure}[t]
	\centering
	\includegraphics[width=\columnwidth]{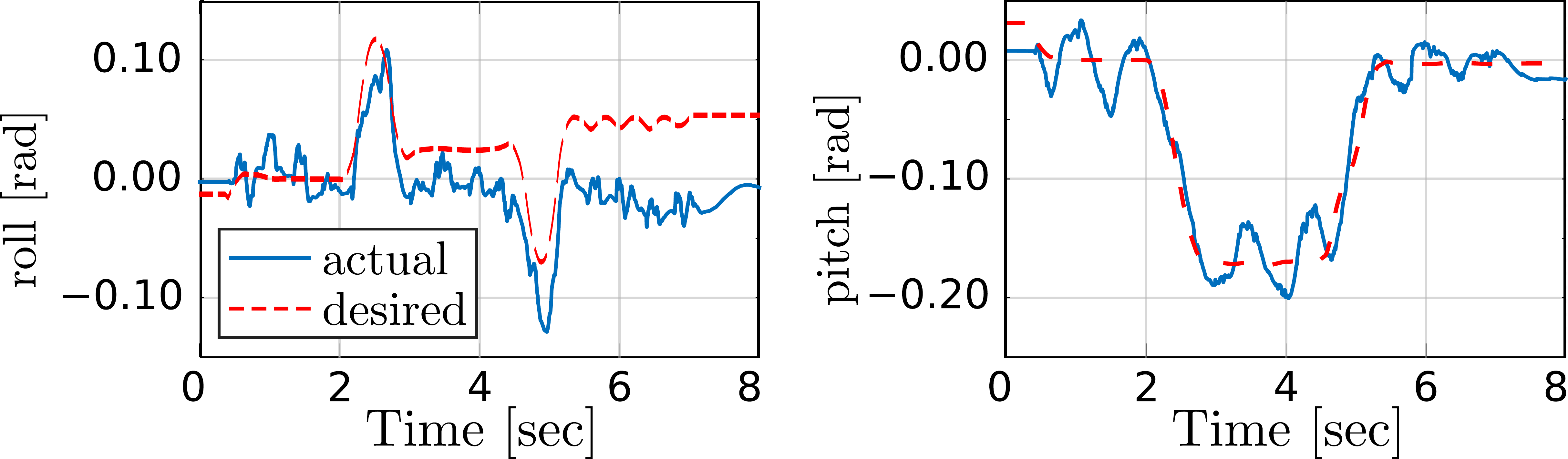}
	\reducespace
	\caption{\small Real experiments. Roll and pitch tracking performance while climbing up a ramp with a
	trotting gait. Although there is under-actuation the controller can still
	track roll and pitch motions.} 
	\label{fig:trottingTracking}
\end{figure}

\textcolor{black}{The quadrupedal trotting gait is} difficult to control because the robot uses only two
legs at the time to achieve the tracking of the desired \gls{com} 
motion and of the trunk orientation. Figure \ref{fig:trottingTracking} depicts the roll and pitch tracking for
climbing up a ramp during a trotting gait. Although a trot is an under-actuated gait, 
our controller can still track the desired orientation. However, in
these cases, the orientation error is always below \unit[0.2]{rad}.

%
\section{Conclusion}\label{sec:conclusion}
\textcolor{black}{This paper presented an experimental validation of our passive \gls{wbc}}.
\textcolor{black}{Compared to our previous work \cite{focchi2017auro}, the presented \gls{wbc}} 
\textcolor{black}{enables higher} dynamic motions thanks to the use of the  full dynamics of the robot. 
Although  similar controllers have been proposed in 
the literature (e.g. \cite{Herzog2016,Henze2016,Bellicoso2018}), 
\textcolor{black}{we validated our locomotion controller  on \gls{hyq} over a wide range of challenging terrain (slopes, 
gaps, stairs, 
etc.), using 
different gaits (crawl and trot).}
%
%
Additionally\textcolor{black}{,} we have analyzed the controller capabilities against 
1) \textcolor{black}{inaccurate friction coefficient estimation},
2) unstable footholds, 
3) changes in the regularization scheme and 
4) \textcolor{black}{the load redistribution under restrictive torque limits.} 
%
%
Extensive experimental results validated the
controller performance together with online \textcolor{black}{terrain} mapping and state estimation.
Moreover, we demonstrated experimentally the superiority of our \gls{wbc} \textcolor{black}{compared}
to a quasi-static control scheme \cite{focchi2017auro}.
%

\footnotesize


\end{document}